\documentclass[10pt,twocolumn,letterpaper]{article}

\usepackage{wacv}
\usepackage{times}
\usepackage{epsfig}
\usepackage{graphicx}
\usepackage{amsmath}
\usepackage{amssymb}
\usepackage{mathtools}
\usepackage{subcaption}

\usepackage{hyperref}
\hypersetup{colorlinks=true,linkcolor=blue,filecolor=magenta,urlcolor=blue}

\wacvfinalcopy % *** Uncomment this line for the final submission

\def\wacvPaperID{****} % *** Enter the wacv Paper ID here

\ifwacvfinal\fi
\begin{document}

%%%%%%%%% TITLE
\title{Depth Completion via Deep Basis Fitting}

% Authors at the same institution
\author{Chao Qu \hspace{2cm} Ty Nguyen \hspace{2cm} Camillo J. Taylor \\
University of Pennsylvania\\
{\tt\small quchao@seas.upenn.edu}
}

\maketitle
\ifwacvfinal\thispagestyle{empty}\fi

%%%%%%%%% ABSTRACT
\begin{abstract}
In this paper we consider the task of image-guided depth completion where our system must infer the depth at every pixel of an input image based on the image content and a sparse set of depth measurements.
We propose a novel approach that builds upon the strengths of modern deep learning techniques and classical optimization algorithms and significantly improves performance. 
The proposed method replaces the final $1\times 1$ convolutional layer employed in most depth completion networks with a least squares fitting module which computes weights by fitting the implicit depth bases to the given sparse depth measurements. 
In addition, we show how our proposed method can be naturally extended to a multi-scale formulation for improved self-supervised training.
We demonstrate through extensive experiments on various datasets that our approach achieves consistent improvements over state-of-the-art baseline methods with small computational overhead.

 %   In this paper, we propose a simple least squares fitting (\textbf{LSF}) module to replace the final $1\times 1$ convolutional layer in various deep neural networks for image-guided depth completion task.
%    Instead of learning the weights of the last layer during training and fix for inference, we compute them directly from the network output and the sparse data given.
%    This is achieved by a least squares fitting from the final feature maps of the decoder to the sparse depths at valid pixel locations.
%    In addition, we show how our proposed method can be naturally extended to a multi-scale formulation for improved self-supervised training.
%    We demonstrate through extensive experiments on various datasets that our approach achieves consistent improvements over a state-of-the-art baseline method with minimal computational overhead.
\end{abstract}

%%%%%%%%% BODY TEXT
\section{Introduction}
Deep convolutional networks have proven to be effective tools for solving deep regression problems like depth prediction and depth completion \cite{Eigen2014DepthMP}. 
Most networks proposed for this regression task share a common structure where the penultimate features are reduced to single channel by a final convolutional layer.
This final convolutional output is then passed through a nonlinear function to map it onto the range of acceptable depth values.

This observation motivates the main contribution of this paper:
Instead of using a fixed set of weights in the final layer, we perform a least squares fit from the penultimate features to the sparse depths to get a set of \emph{data-dependent} weights.
The rest of the network parameters are still shared across input data and learned using stochastic gradient descent.
From a regression point of view, the network that produces the penultimate layer of features is an adaptive basis function \cite{Bishop:2006:PRM:1162264} and we refer to the features before the final layer as \textit{depth bases}.
We argue that explicitly carrying out a regression from the depth bases to the sparse depths allows the network to learn a different representation that better enforce its predictions to be consistent with the measurements, which manifests as significant performance gain.

To this end, we first demonstrate how one could circumvent the nonlinearity from the depth activation function by solving a linear least squares problem with transformed target sparse depths. 
We then address the full robustified nonlinear least squares problem in order to deal with noisy measurements and outliers in real-world data.
Finally, to make our module truly a drop-in replacement for the final convolutional layer, we show how to adapt it to output predictions at multiple scales with progressively increased detail, which is a feature required by self-supervised training schemes.

\section{Related Work}

\subsection{Depth Estimation}

\noindent
\textbf{Supervised Learning.}
Estimating dense depths from a single image is a fundamentally ill-posed problem.
Recent learning-based approaches try to solve this by leveraging the predictive power of deep convolutional neural networks (CNN) with strong regularization \cite{Eigen2014DepthMP, Laina2016DeeperDP, Fu2018DeepOR}.
These works require dense or semi-dense ground truth annotations, which are costly to obtain in large quantities in practice.
Synthetic data \cite{Qiu2016UnrealCVCC, Gaidon2016VirtualWorldsAP, Ros2016TheSD}, on the other hand, can be generated more easily from current graphics systems.
However, it is non-trivial to generate synthetic data that closely matches the appearance and structure of the real-world, thus the resulting networks may require an extra step of fine-tuning or domain adaptation \cite{Abarghouei2018RealTimeMD}.

\vspace{\baselineskip}
\noindent
\textbf{Self-Supervised Learning.}
When ground truth depths are not available, one could instead seek to use view synthesis as a supervisory signal \cite{Szeliski1999PredictionEA}.
This so-called self-supervised training has gained popularity in recent years \cite{Mahjourian2018UnsupervisedLO, Pillai2018SuperDepthSS, Wang2017LearningDF}.
The network still takes a single image as input and predicts depths, but the loss is computed on a set of images.
This is achieved by warping pixels from a set of source images to the target image using the predicted depths, along with estimated camera poses and camera intrinsics.
Under various constancy assumptions \cite{Papenberg2005HighlyAO}, errors between target and synthesized images are computed and back-propagated through the network for learning.

Another version of self-supervision utilizes synchronized stereo pairs \cite{Garg2016UnsupervisedCF} during training.
In this setting, the network predicts the depth for the left view and uses the known focal length and baseline to reconstruct the right view, and vice versa.
A more complex form utilizes the motion in monocular videos \cite{Zhou2017UnsupervisedLO}. In these approaches the network also needs to predict the transformation between frames.
The biggest challenge faced by monocular self-supervision is handling moving objects.
Many authors try to address this issue by predicting an explanability mask \cite{Zhou2017UnsupervisedLO}, motion segmentation \cite{Vijayanarasimhan2017SfMNetLO},
and joint optical-flow estimation \cite{Yin2018GeoNetUL}.
We refer readers to \cite{Godard2018DiggingIS} for a more detailed review.

\subsection{Depth Completion}
Depth completion is an extension to the depth estimation task where sparse depths are available as input.
Uhrig \etal \cite{Uhrig2017SparsityIC} propose a sparse convolution layer that explicitly handles missing data, which renders it invariant to different levels of sparsity.
Ma \etal \cite{Ma2018SelfsupervisedSS} adopt an early-fusion strategy to combine color and sparse depths inputs in a self-supervised training framework.
On the other hand, Jaritz \etal \cite{Jaritz2018SparseAD} and Shivakumar \etal \cite{Shivakumar2019DFuseNetDF} advocate a late-fusion approach to transform both inputs into a common feature space.
Zhang \etal \cite{Zhang2018DeepDC} and Qiu \etal \cite{Qiu2018DeepLiDARDS} estimate surface normals as a secondary task to help densify the sparse depths.
Irman \etal \cite{Imran2019DepthCF} identify the cause of artifacts caused by convolution on sparse data and propose a novel scheme, Depth Coefficients, to address this problem.
Eldesokey \etal \cite{Eldesokey2018PropagatingCT} and Gansbeke \cite{Gansbeke2019SparseAN} propose to use a confidence mask to handle noise and uncertainty in sparse data.
Yang \etal \cite{Yang2019DenseDP} infer the posterior distribution of depth given an image and sparse depths by a Conditional Prior Network.
While most of the above works deal with data from LiDARs or depth cameras, Wong \etal \cite{Wong2019VOICEDDC} design a system that works with very sparse data from a visual-inertial odometry system.
Weeraskera \etal \cite{Weerasekera2018JustinTimeRI} attach a fully-connected Conditional Random Field to the output of a depth prediction network, which can also handle any input sparsity pattern.

Cheng \etal \cite{Cheng2018DepthEV} propose a convolutional spatial propagation network that learns the affinity matrix to complete sparse depths. This is similar to a diffusion process and uses several iterations to update the depth map.
Another iterative approach is described by Wang \etal \cite{wang2018pnp}, in which they design a module that can be integrated into many existing methods to improve performance of a pre-trained network without re-training.
This is done by iteratively updating the intermediate feature map to make the model output consistent with the given sparse depths.
% However, they need to perform one forward pass on each iteration, which results in a larger overhead at inference time.
% Our approach requires the network to be retrained, but only adds a small amount of computation time during inference, which is more suitable for resource constraint platforms.
Like \cite{wang2018pnp}, our approach could be readily integrated into many of the previously proposed depth completion networks.

In other related work Tang \etal \cite{tang2018banet}. 
propose to parameterize depth map with a set of basis depth maps and optimize weights to minimize a feature-metric distance. 
In contrast, our bases are multi-scale by construction and are fit directly to the sparse depths.

\section{Method}

In this section, we describe our proposed method for the task of monocular image-guided depth completion\footnote{From now on we will refer to this task as depth completion.}.
Given an image $X$ and a sparse depth map $S$, we wish to predict a dense depth image $D^\prime$ from a depth estimation function $f$ that minimizes some loss function $\mathcal{L}$ with respect to the ground truth depth $D$.
Typically, $X$ is a color image, $S$ the sparse depth map where invalid pixels are encoded by $0$, and $f$ a fully convolutional neural network whose parameters are denoted by $\theta$. 
When ground-truth depth $D$ is available, the learning problem is to determine $\theta^*$ according to
\begin{equation}
    \theta^* = \arg\min_\theta \mathcal{L}(f(X, S; \theta), D)
\end{equation}
For supervised training we choose $\mathcal{L}$ to be the L1 norm on depth and for self-supervised training we use a combination of L1+SSIM on the intensity values \cite{Wang2004ImageQA} coupled with an edge-aware smoothness term \cite{Godard2018DiggingIS}. 
\subsection{Linear Least-Squares Fitting (LSF) Module}\label{sec:lsf}

\begin{figure}[t]
    \begin{center}
        \includegraphics[width=\linewidth]{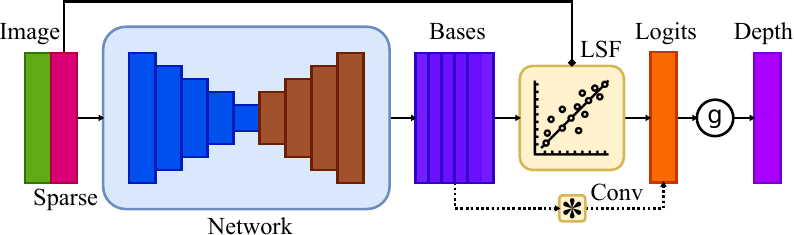}
    \end{center}
    \caption{An overview of our proposed method. Solid lines indicate the data flow of our module, while dotted lines indicate that of the baseline method, which is simply a convolutional layer. 
    Our LSF module can replace the convolutional layer with no change to the rest of the network.}
    \label{fig:method/overview}
\end{figure}

Existing depth prediction networks usually employ a final convolutional layer to convert an $M$-channel set of basis features, $B$, to a single-channel result, $L$, which is sometimes referred to as the logits layer. The inputs to this final layer are allowed to range freely between $-\infty$ and $+\infty$ while the logit outputs are mapped to positive depth values by a nonlinear activation function, $g$.
Following common practice in the depth completion literature \cite{Godard2018DiggingIS} we choose $g$ as follows:
\begin{equation}
    g(x) = a / \sigma(x) = a(1 + e^{-x})
\end{equation}
where $a$ is a scaling factor that controls the minimum depth and $\sigma(\cdot)$ the sigmoid function. In this work, we set $a=1$.

For simplicity we assume that the final convolution filter that maps the basis features, $B$, onto the logits, $L$, has a kernel size of $1 \times 1$ with bias $w_0$, but one could easily extend our result to arbitrary kernel size.
$L$ is, therefore, an affine combination of channels in $B$ and the predicted depth at pixel $i$ is
\begin{equation}
    D^\prime[i] = g\left(L[i]\right) 
    = g\left(\sum_{j=0}^{M} w_j \cdot B_j[i] \right) 
    = g(\mathbf{w}^\top \mathbf{b}_i)
\end{equation}
where $\mathbf{w}=(w_0, \cdots, w_{M})^\top$ represents the combined filter weights and bias, and $\mathbf{b}_i$ the basis (feature) vector at pixel $i$ with $B_0[i]=1$, and $[\cdot]$ the pixel index operator.
To simplify notations, we use lower case letters, \eg $\mathbf{b}_i = B[i]$, to denote values at a particular pixel location.
The weights $\mathbf{w}$ are updated via \textit{back-propagation} \cite{LeCunBP1988} using stochastic gradient descent \cite{Bottou2010LargeScaleML}. Once learned they are typically fixed at inference time.

When enough sparse depth measurements are available the weights $\mathbf{w}$ can instead be directly computed from data.
Specifically, our weights are obtained through a least squares fit from the bases $B$ to the sparse depths $S$ at valid pixels, which can then be used in place of the final convolutional layer.
An overview of our proposed method is shown in Figure \ref{fig:method/overview}.

The objective function we wish to minimize for the least squares problem is
\begin{equation}\label{eqn:obj}
    \min_\mathbf{w} \frac{1}{2}
    \sum_{i=1}^N r^2 
    \left( \mathbf{w}, \mathbf{b}_i, s_i \right)
\end{equation}
with residual function
\begin{equation}\label{eqn:res_nonlin}
    r(\mathbf{w}, \mathbf{b}_i, s_i)
    = g\left(\sum_{j=0}^M \mathbf{w}_j \mathbf{b}_{ij}\right) - s_i
    = g\left(\mathbf{w}^\top \mathbf{b}_i \right) - s_i
\end{equation}
where $s_i$ denotes an individual sparse depth measurement, $N$ is the number of valid pixels in $S$, $M$ the number of channels in $B$,
and $g(\cdot)$ a nonlinear activation function.

The residual function $r(\cdot)$ is obviously nonlinear \wrt the weights $\mathbf{w}$ due to the nonlinearity in $g(\cdot)$. 
A simple workaround is to transform the target variable $s$ by $g^{-1}(\cdot)$ to arrive at a new linear residual function
\begin{equation}
    \Tilde{r}(\mathbf{w}, \mathbf{b}_i, s_i)
    = \mathbf{w}^\top\mathbf{b}_i - g^{-1}(s_i)
    = \mathbf{w}^\top\mathbf{b}_i - t_i
\end{equation}
We can then rewrite the new objective function (\ref{eqn:obj}) in matrix form to obtain a linear least squares problem
\begin{equation} \label{eqn:obj_lin}
    \min_\mathbf{w} 
    \frac{1}{2} \left\| \mathbf{B} \mathbf{w} -\mathbf{t} \right\|^2
\end{equation}
where $\mathbf{B}$ denotes the $N\times (M+1)$ matrix of stacked features $\mathbf{b}_i$ at valid pixel locations and $\mathbf{t}$ the corresponding transformed sparse depths vector. The solution to (\ref{eqn:obj_lin}) is the well-known \textit{Moore-Penrose pseudo-inverse} which can be further regularized with parameter $\lambda$ \cite{Bishop:2006:PRM:1162264}.
\begin{equation}
    \mathbf{w}^* 
    = \left(\lambda \mathbf{I} +  \mathbf{B}^\top \mathbf{B}\right)^{-1} 
    \mathbf{B}^\top \mathbf{t}
\end{equation}
Notice here that our weights $\mathbf{w}^*$ are calculated deterministically as a function of the bases $B$ and the sparse depth $S$, while the original convolution filter is independent of both.
In practice, this problem is usually solved via LU or Cholesky decomposition
both of which are differentiable \cite{Murray2016DifferentiationOT}.
Thus, the entire training process including our LSF module is differentiable which means that it can be trained in an end-to-end manner. 
This is an important point since we have found that retraining the network with this fitting module produces much better results than simply adding the fitting procedure to a pretrained network without retraining.
Effectively the retraining allows the network to make best use of the new adaptive fitting layer.
\subsection{Robustified Nonlinear Fitting}

The linear LSF module is readily usable as a replacement for the final convolution layer in many depth prediction networks.
One problem remains to be addressed, which is the fact that the original objective function in Equation \ref{eqn:res_nonlin} is nonlinear \wrt the weights $\mathbf{w}$.
Although applying the inverse transformation $g^{-1}(\cdot)$ to the sparse depths is a simple yet effective solution, we show that performing a full robustified nonlinear least squares fitting provides further performance improvements and outlier rejection at the cost of extra computation time.

Real-world data often contain noise and outliers that are hard to model or eliminate.
Cheng \etal \cite{Cheng2019NoiseAwareUD} point out that there exist 
many different types of noise in LiDAR data from the well-known KITTI \cite{Geiger2013IJRR} dataset.
These include:
1) noise in the LiDAR measurement itself,
2) LiDAR camera misalignment,
3) moving objects,
and 4) transparent and reflective surfaces.
They propose a novel feedback loop that utilizes stereo matching from the network to clean erroneous data points in the sparse depths. 
Gansbeke \etal \cite{Gansbeke2019SparseAN} let the network predict a confidence map to weight information from different input branches. 
% These methods, although demonstrated to be useful, add a fair amount of parameters and computation overhead to the network. 
To handle these cases, we employ M-estimators \cite{Huber.Wiley.ea1981Robuststatistics}, which fit well within our least squares framework.

Recall the objective function in equation (\ref{eqn:obj}), taking the derivative with respect to $\mathbf{w}$, setting it to zero and ignoring higher-order terms yields the following linear equation (Gauss-Newton approximation)
\begin{equation}
    \mathbf{J}^\top  \mathbf{J} \Delta \mathbf{w} 
    = -\mathbf{J}^\top \mathbf{r}
\end{equation}
where $\mathbf{J}$ is the Jacobian matrix that is formed by stacking Jacobians $\mathbf{J}_i(\mathbf{w}, \mathbf{b}_i, s_i) = \partial r(\mathbf{w}, \mathbf{b}_i, s_i)/\partial \mathbf{w}$, 
and $\mathbf{r}$ is the residual vector formed by stacking $r_i(\mathbf{w}, \mathbf{b}_i, s_i)$.
Following standard practice in Triggs \etal \cite{Triggs1999BundleA}, we minimize the \textbf{effective squared error} where the cost function is statistically weighted and robustified, which is equivalent to solving for $\Delta{\mathbf{w}}$ in
\begin{equation}\label{eqn:weighted_normal}
    \Bar{\mathbf{J}}^\top \mathbf{W} \Bar{\mathbf{J}} \Delta \mathbf{w}
    = -\Bar{\mathbf{J}}^\top \mathbf{W} \Bar{\mathbf{r}}
\end{equation}
\begin{equation}
    \text{with} \quad
    \Bar{\mathbf{J}}_i = \sqrt{\rho^\prime_i}  \mathbf{J}_i
    \quad \text{and} \quad
    \Bar{\mathbf{r}}_i = \sqrt{\rho^\prime_i}  \mathbf{r}_i
\end{equation}
where $\mathbf{W}=\mathbf{L}^\top \mathbf{L}$ a diagonal matrix with terms inverse-proportional to the noise in each measurement, which we assume to be Gaussian for LiDARs , $\rho(x)$ is the Huber loss \cite{huber1964} and $\rho^\prime$ its first derivative
\begin{equation}
    \rho(x) = 
    \begin{cases}
        x^2, & |x| \leq 1 \\
        2|x| - 1, & |x| > 1
    \end{cases}
\end{equation}
We iteratively calculate $\Delta\mathbf{w}$ by solving (\ref{eqn:weighted_normal}) and update $\mathbf{w}$ 
\begin{equation}
    \mathbf{w} \leftarrow \mathbf{w} + \Delta \mathbf{w}
\end{equation}
with $\mathbf{w}$ initialized from the linear fitting in Section \ref{sec:lsf}.

Theoretically, one should repeat this until convergence, but to alleviate the problem of vanishing or exploding gradients \cite{Hochreiter:01book}, 
we adopt the fixed-iteration approach used in \cite{tang2018banet}, which is also known as an \textit{incomplete optimization} \cite{pmlr-v22-domke12}.
Despite its limitations, it has the advantage of having a fixed training/inference time and reduced memory consumption, which is often desirable in robotic systems with limited computational resources.
As discussed in earlier Section \ref{sec:lsf}, solving a linear system like equation (\ref{eqn:weighted_normal}) via Cholesky decomposition is differentiable, thus optimizing this nonlinear objective function by performing a fixed number of Gauss-Newton steps maintains the differentiability of the entire system.
\subsection{Multi-scale Prediction for Self-supervision}
\label{sec:self}

Self-supervised training formulates the learning problem as novel view synthesis, where the network predicted depth is used to synthesize a target image from other viewpoints. 
To overcome the gradient locality problem of the bilinear sampler \cite{Jaderberg2015SpatialTN} during image warping, 
previous works \cite{Godard2016UnsupervisedMD, Zhou2017UnsupervisedLO} adopt a multi-scale prediction and image reconstruction scheme by predicting a depth map at each decoder layer's resolution.
According to Godard \etal \cite{Godard2018DiggingIS}, this has the side effect of creating artifacts in large texture-less regions in the lower resolution depth maps due to ambiguities in photometric errors.
They later improved upon this multi-scale formulation by upsampling all the lower resolution depth maps to the input image resolution. 

This technique greatly reduces various artifacts in the final depth prediction, 
but it still has one undesired property, namely, depth maps predicted at each scale are largely independent. 
Lower resolution depth maps are used in training phase, 
but are discarded during inference, resulting in a waste of parameters.

\begin{figure}[h!]
    \begin{center}
        \includegraphics[width=0.85\linewidth]{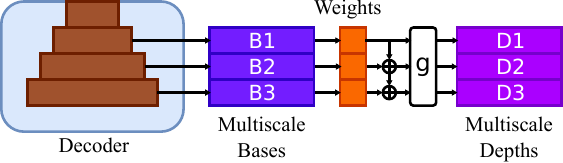}
    \end{center}
    \caption{Our proposed multi-scale depth prediction. 
    The full resolution depth $D_3$ is reconstructed using all bases prediction.}
    \label{fig:method/multiscale}
\end{figure}

Rather than predicting a depth map $D^{\prime}_k$ at each scale $k$ separately, we propose to predict a set of bases $B_k$, as shown in Figure \ref{fig:method/multiscale}. 
Each of the basis vectors is obtained by upsampling features from corresponding scales in the decoder as shown in Figure \ref{fig:method/multiscale} so the resulting basis images are \emph{band-limited} by construction with coarser basis images corresponding to earlier layers in the decoder.
The depth prediction at a particular scale $s$ is then reconstructed using bases up to that scale. 
\begin{equation}
    d_{is}^{\prime}
    = g\left(\sum_{k=0}^s \mathbf{w}_k^{\top} \mathbf{b}_{ik}  \right)
\end{equation}
The final depth prediction at highest scale $K$ is 
\begin{equation}
    d^\prime_i \coloneqq d_{iK}^\prime 
    = g\left(\sum_{k=0}^K \mathbf{w}_k^\top \mathbf{b}_{ik}\right)
    = g\left(\mathbf{w}^\top \mathbf{b}_i \right)
\end{equation}
where $\mathbf{b}_i = (\mathbf{b}_{i0}^\top, \ldots, \mathbf{b}_{iK}^\top)^\top$ 
and $\mathbf{w}=(\mathbf{w}_0^\top, \ldots, \mathbf{w}_K^\top)^\top$. 

With this formulation, predictions at different scales will work towards the same goal, which is to reconstruct the full resolution depth map. 
This approach is analogous to wavelet or Fourier encodings of an image where the basis maps are organized into band-limited components to represent the signal at various scales.

Our LSF module handles this multi-scale approach quite naturally since we can simply allocate the basis maps in $B$ amongst the desired scales, then upsample and group them back together to perform the fitting step.
Henceforth we use this new multi-scale prediction scheme in all our experiments, even for supervised training where only the full resolution depth prediction is required.

\begin{table}
\begin{center}\small
\begin{tabular}{|l|c|c|c|c|}
\hline
Dataset & Resolution        & \# Train & \# Val & Cap [m] \\
\hline\hline
KITTI \cite{Geiger2013IJRR, Uhrig2017SparsityIC} & 375 $\times$ 1242 & 38412    & 3347   & 80      \\
V-KITTI \cite{Gaidon2016VirtualWorldsAP}         & 188 $\times$ 621  & 5156     & 837    & 130     \\
Synthia \cite{Ros2016TheSD}                      & 304 $\times$ 512  & 3634     & 901    & 130     \\
NYU-V2 \cite{Silberman2012IndoorSA}              & 480 $\times$ 640  & 1086     & 363    & -       \\
\hline
\end{tabular}
\end{center}
\caption{A summary of all datasets used.
\textbf{Cap} indicates the maximum depth being used for sampling sparse depths as well as in computing various error metrics.
\textbf{Resolution} is the image resolution that we use in our experiments, which we downsample from the original one if necessary.}
\label{tab:data}
\end{table}

\section{Experiments}

\begin{table*}[h!]
\begin{center}\footnotesize
\begin{tabular}{|c|c|c|c|c|c|c|c|c|c|c|c|c|c|c|}
\hline
\multicolumn{3}{|c|}{Supervised Training} & \multicolumn{3}{c|}{NYU-V2} & \multicolumn{3}{c|}{V-KITTI} & \multicolumn{3}{c|}{Synthia} & \multicolumn{3}{c|}{KITTI}\\ \hline
Input & Method & Sparse & MAE & RMSE & $\delta_1$ & MAE & RMSE & $\delta_1$ & MAE & RMSE & $\delta_1$ & MAE & RMSE & $\delta_1$ \\ \hline\hline
rgb & conv & - & 0.6244 & 0.8693 & 58.44 & 6.9998 & 14.653 & 66.43 & 2.3911 & 6.3915 & 76.09 & 1.8915 & 4.1164 & 86.24 \\
rgb & pnp & 0.2\% & 0.5517 & 0.7976 & 64.23 & 6.4701 & 13.990 & 70.18 & 2.1716 & 6.0084 & 81.37 & 1.6581 & 3.8019 & 88.67 \\
rgb & lsf- & 0.2\% & 0.4081 & 0.6124 & 77.86 & 5.8379 & 12.712 & 71.62 & 2.4089 & 6.2520 & 78.49 & 1.7033 & 3.5986 & 91.80\\
rgb & lsf & 0.2\% & \textbf{0.1826} & \textbf{0.3165} & \textbf{96.11} & \textbf{4.5122} & \textbf{9.7933} & \textbf{77.18} & \textbf{2.0104} & \textbf{5.6285} & \textbf{84.37} & \textbf{0.7716} & \textbf{2.0808} & \textbf{97.69} \\ \hline
\multicolumn{3}{|c|}{(conv-lsf) / conv} & +71\% & +64\% &  & +36\% & +33\% & & +16\% & +12\% & & +59\% & +50\% & \\ \hline\hline
rgbd & conv & 4\% & 0.1089 & 0.1679 & 99.20 & 1.5683 & 4.8982 & 94.71 & 0.7506 & 3.3322 & 96.50 & 0.3033 & 1.1392 & 99.57 \\
rgbd & pnp & 4\% & 0.1008 & 0.1604 & 99.24 & 1.5301 & 4.8798 & 94.81 & 0.7311 & 3.3217 & 96.60 & 0.2993 & 1.1343 & 99.57 \\
rgbd & lsf- & 4\% & 0.1127 & 0.1853 & 99.34 & 2.1049 & 6.1901 & 95.30 & 1.3220 & 4.6594 & 94.27 & 0.6319 & 2.2895 & 98.46 \\
rgbd & lsf & 4\% & \textbf{0.0300} & \textbf{0.0735} & \textbf{99.83} & \textbf{1.2598} & \textbf{4.6227} & \textbf{97.43} & \textbf{0.5317} & \textbf{3.1146} & \textbf{97.85} & \textbf{0.2266} & \textbf{0.9988} & \textbf{99.67} \\ \hline
\multicolumn{3}{|c|}{(conv-lsf) / conv} & +72\% & +56\% &  & +20\% & +6\% & & +29\% & +7\% & & +25\% & +12\% & \\ \hline
\end{tabular}
\end{center}
\caption{Quantitative results of supervised training on various datasets.
\textbf{conv} denotes the baseline network, 
\textbf{pnp} denotes running the PnP \cite{wang2018pnp} module on the trained \textbf{conv} network without re-training,
\textbf{lsf-} indicates adding a linear LSF module to the pre-trained \textbf{conv} network without re-training for 5 iterations, and
\textbf{lsf} is our linear fitting module (re-trained).
Percentage values listed under the \textbf{Sparse} column indicates sparse depths percentage of image resolution.
Best results in each category are in \textbf{bold}.}
\label{tab:exp/perf_sup}
\end{table*}

\begin{table*}[h!]
\begin{center}\footnotesize
\begin{tabular}{|c|c|c|c|c|c|c|c|c|c|c|c|c|c|c|}
\hline
\multicolumn{3}{|c|}{Noise and Outliers} & \multicolumn{3}{c|}{NYU-V2} & \multicolumn{3}{c|}{V-KITTI} & \multicolumn{3}{c|}{Synthia} & \multicolumn{3}{c|}{KITTI} \\ \hline
Input & Method & Sparse & MAE & RMSE & $\delta_1$ & MAE & RMSE & $\delta_1$ & MAE & RMSE & $\delta_1$ & MAE & RMSE & $\delta_1$ \\ \hline\hline
rgb & pnp & 0.2\% & 0.5587 & 0.8019 & 63.66 & 6.5099 & 14.018 & 69.86 & 2.2044 & 6.0268 & 80.89 & 1.6571 & 3.8019 & 88.67 \\
rgb & lsf & 0.2\% & 0.2439 & 0.3815 & 92.93 & 5.2670 & 10.696 & 65.00 & 2.2197 & 5.9136 & 78.34 & 0.7716 & 2.0808 & 97.69 \\
rgb & lsf2 & 0.2\% & 0.2304 & 0.3519 & 92.70 & 6.0025 & 10.768 & 51.01 & 3.2160 & 7.2096 & 59.68 & 1.0111 & 2.4547 & 95.88 \\
rgb & lsf2+ & 0.2\% & \textbf{0.1880} & \textbf{0.3217} & \textbf{94.97} & \textbf{4.6786} & \textbf{9.7402} & \textbf{70.16} & \textbf{2.1032} & \textbf{5.7685} & \textbf{79.00} & \textbf{0.6775} & \textbf{1.9651} & \textbf{98.28} \\ \hline 
\multicolumn{3}{|c|}{(lsf - lsf2+) / lsf}  & +23\% & +16\% & & +11\% & +9\%  & & +5\% & +2\% & & +12\% & +6\% & \\ \hline \hline
rgbd & conv & 4\% & 0.1173 & 0.1788 & 99.07 & 1.8748 & 5.1880 & 94.17 & 0.8774 & 3.4660 & 96.03 & 0.3033 & 1.1392 & 99.57 \\
rgbd & pnp & 4\% & 0.1061 & 0.1688 & 99.15 & 1.8067 & 5.1342 & 94.46 & 0.8452 & 3.4511 & 96.19 & 0.2993 & 1.1343 & 99.57 \\
rgbd & lsf & 4\% & 0.0606 & 0.1102 & 99.73 & 1.8599 & 5.1987 & \textbf{95.90} & \textbf{0.7082} & \textbf{3.2426} & \textbf{97.41} & 0.2266 & 0.9988 & 99.67 \\
rgbd & lsf2 & 4\% & 0.0577 & 0.1080 & 99.72 & 1.8008 & \textbf{5.0008} & 94.58 & 0.7890 & 3.4142 & 96.78 & 0.2305 & 1.0417 & 99.67 \\
rgbd & lsf2+ & 4\% & \textbf{0.0493} & \textbf{0.1003} & \textbf{99.73} & \textbf{1.7273} & 5.0422 & 95.50 & 0.7188 & 3.2579 & 97.31 & \textbf{0.2208} & \textbf{0.9758} & \textbf{99.71} \\ \hline
\multicolumn{3}{|c|}{(conv - lsf2+) / conv}  & +58\% & +44\% & & +8\% & +3\% & & +18\% & +6\% & & +27\% & +14\% & \\ \hline
\end{tabular}
\end{center}
\caption{Quantitative results of supervised training with noisy data and outliers.
For all datasets except KITTI,
noise is additive Gaussian with standard deviation of 0.05m.
We randomly sample 30\% of sparse depths to be outliers.
\textbf{conv} denotes the baseline network, 
\textbf{pnp} denotes running the PnP \cite{wang2018pnp} module on the trained \textbf{conv} network without re-training,
\textbf{lsf} is our linear fitting module, 
\textbf{lsf2} is our nonlinear fitting module with 2 iterations, 
and \textbf{lsf2+} is \textbf{lsf2} with robust norm (Huber).
Best results in each category are in \textbf{bold}.}
\label{tab:exp/perf_noise}
\end{table*}

\begin{table*}[h!]
\begin{center}\footnotesize
\begin{tabular}{|c|c|c|c|c|c|c|c|c|c|c|c|c|c|c|}
\hline
\multicolumn{3}{|c|}{Self-Supervised Training} & \multicolumn{3}{c|}{V-KITTI Mono} & \multicolumn{3}{c|}{Synthia Mono} & \multicolumn{3}{c|}{Synthia Stereo} & \multicolumn{3}{c|}{KITTI Stereo} \\ \hline
Input & Method & Sparse & MAE & RMSE & $\delta_1$ & MAE & RMSE & $\delta_1$ & MAE & RMSE & $\delta_1$ & MAE & RMSE & $\delta_1$ \\ \hline\hline
rgbd & conv-ms & 4\% & 2.9904 & 7.4517 & 86.87 & 3.0191 & 9.1076 & 66.43 & 1.3498 & 5.8643 & 92.73 & 0.6295 & 2.0950 & \textbf{99.00} \\
rgbd & lsf & 4\% & \textbf{2.3804} & \textbf{6.7326} & \textbf{93.76} & \textbf{1.4564} & \textbf{4.6260} & \textbf{91.76} & \textbf{0.8619} & \textbf{3.9523} & \textbf{96.30} & \textbf{0.5820} & \textbf{1.7370} & 98.79 \\ \hline
\multicolumn{3}{|c|}{(conv-ms - lsf) / conv-ms} & +20\% & +10\%  & & +52\% & +49\% & & +36\% & +33\% & & +8\% & +17\% & \\ \hline
\end{tabular}
\end{center}
\caption{Quantitative results of self-supervised training on various datasets.
The densely labeled NYU-V2 is random and monocular,
thus is excluded from this experiment.
Here \textbf{conv-ms} is the baseline multi-scale prediction,
\textbf{lsf} is the our proposed method with linear fitting and multi-scale basis.
Best results in each category are in \textbf{bold}.}
\label{tab:exp/perf_self}
\end{table*}

\subsection{Implementation Details}

% ==============================================================================
\noindent \textbf{Network Architecture.}
All networks and training are implemented in PyTorch \footnote {\url{http://pytorch.org}}.
To investigate the effectiveness of the proposed LSF module, we adopt the network used in Ma \etal \cite{Ma2018SelfsupervisedSS} as our main baseline.  
The network is a symmetric encoder-decoder \cite{RFB15a} with skip connections.
We make the following modifications for better training:
1) transposed convolutions are replaced with resize convolutions \cite{odena2016deconvolution} for better upsampling,
2) the extra convolution layer between the encoder and the decoder are removed,
3) the encoder is based on ResNet18, as opposed to ResNet34 \cite{He2015DeepRL} and is initialized with parameters pretrained on ImageNet \cite{Russakovsky2014ImageNetLS}.

We let the decoder output 4, 8, 16, and 32-dimensional bases at each scale.
These are then upsampled to the image resolution and concatenated together to form a 60-dimensional basis.
For the baseline network, it is fed directly into a final convolution layer while for ours, it is passed into the LSF module together with the sparse depths.
Therefore, these two methods are exactly the same in terms of network parameters, up to the last convolution layer.

% ==============================================================================
\vspace{\baselineskip}
\noindent \textbf{Training Parameters.} \label{sec:train}
Following \cite{Ma2018SelfsupervisedSS}, we use the Adam optimizer \cite{Kingma2015AdamAM} with an initial learning rate of 1e-4 and reduce it by half every 5 epochs.
Training is carried out on a single Tesla V100 GPU with 15 epochs and the best validation result is reported.
Batch sizes may vary across datasets due to GPU memory constraints, but are kept the same for experiments of the same dataset.
Only random horizontal flips are used to augment the data for supervised training, no data augmentation is performed for self-supervised training.
The above settings are used across \textbf{all} experiments in this work (unless explicitly stated) with the same random seed to ensure controlled experiments with fair and meaningful comparisons.

% ==============================================================================
% ==============================================================================
\subsection{Datasets}
A summary of all datasets we evaluate on is shown in Table \ref{tab:data}.

% ==============================================================================
\vspace{\baselineskip}
\noindent \textbf{KITTI Depth Completion.}
We evaluate on the newly introduced KITTI depth completion dataset \cite{Uhrig2017SparsityIC} and follow the official training/validation split.
The ground truth depth is generated by merging several consecutive LiDAR scans around a given frame and refined using a stereo matching algorithm.
The sparse depth map is generated by projecting LiDAR measurements onto the closest image, which occupies on average 4\% of the image resolution.
We use all  categories from the KITTI raw dataset \cite{Geiger2013IJRR} except for \textbf{Person} as it contains mostly static scenes with moving objects, which is not suitable for self-supervised training.

% ==============================================================================
\vspace{\baselineskip}
\noindent
\textbf{Virtual KITTI.}
The Virtual KITTI (V-KITTI) dataset is a synthetic video dataset \cite{Gaidon2016VirtualWorldsAP}, which contains 50 monocular videos generated with various simulated lighting and weather conditions with dense ground truth annotations.
We adopt an out-of-distribution testing scheme for this dataset.
Specifically, we use sequences 1, 2, 6, 18 with variations \textbf{clone}, \textbf{morning}, \textbf{overcast} and \textbf{sunset} for training, and sequence 20 with variation \textbf{clone} for validation.
Thus the testing sequence is never seen during training.
The sparse depths are generated by randomly sampling pixels that have a depth value less than 130 meters.
We intentionally increase the depth cap to 130 meters for all synthetic datasets since recent LiDAR units \footnote{\url{https://www.ouster.io/}} can easily achieve this range.

% ==============================================================================
\vspace{\baselineskip}
\noindent
\textbf{Synthia.}
Synthia \cite{Ros2016TheSD} is another synthetic dataset in urban settings with dense ground truth.
We use the \textbf{SYNTHIA-Seqs} version which simulates four video sequences acquired from a virtual car across different seasons.
Following the training protocol in V-KITTI, we use sequences 1,2,5,6 for training and sequence 4 for validation, all under the \textbf{summer} variation.
We include this dataset because it has simulated stereo images, which serves as a complement to the monocular only V-KITTI.
Again ground truth and sparse depths are capped at 130 meters.

% ==============================================================================
\vspace{\baselineskip}
\noindent
\textbf{NYU Depth V2.}
In addition to all the outdoor datasets, we also validate our approach on NYU Depth V2 (NYU-V2) \cite{Silberman2012IndoorSA}, which is an indoor dataset.
We use the 1449 densely labeled pairs of aligned RGB and depth images instead of the full dataset which is comprised of raw image and depth data as provided by the Kinect sensor.
The dataset is split into approximately 75\% training and 25\% validation. 
We use the same strategy as above for sampling sparse depths but put no cap on the maximum depth.
\subsection{Results}

We evaluate performance using standard metrics in the depth estimation literature.
Note that for accuracy ($\delta$ threshold) \cite{Eigen2014DepthMP} we only report $\delta_1 < 1.25$,
due to space limitations and the fact that the $\delta_2$ and $\delta_3$ are typically 99\% for our experiments and thus provide limited insights.
Following \cite{wang2018pnp}, we group results based on input modalities, 
where \textbf{rgb} denotes a network that only takes a color image as input.
In contrast \textbf{rgbd} indicates a network that takes both the color image and the sparse depths as inputs.

\begin{figure*}[h!]
\begin{center}
\includegraphics[width=\linewidth]{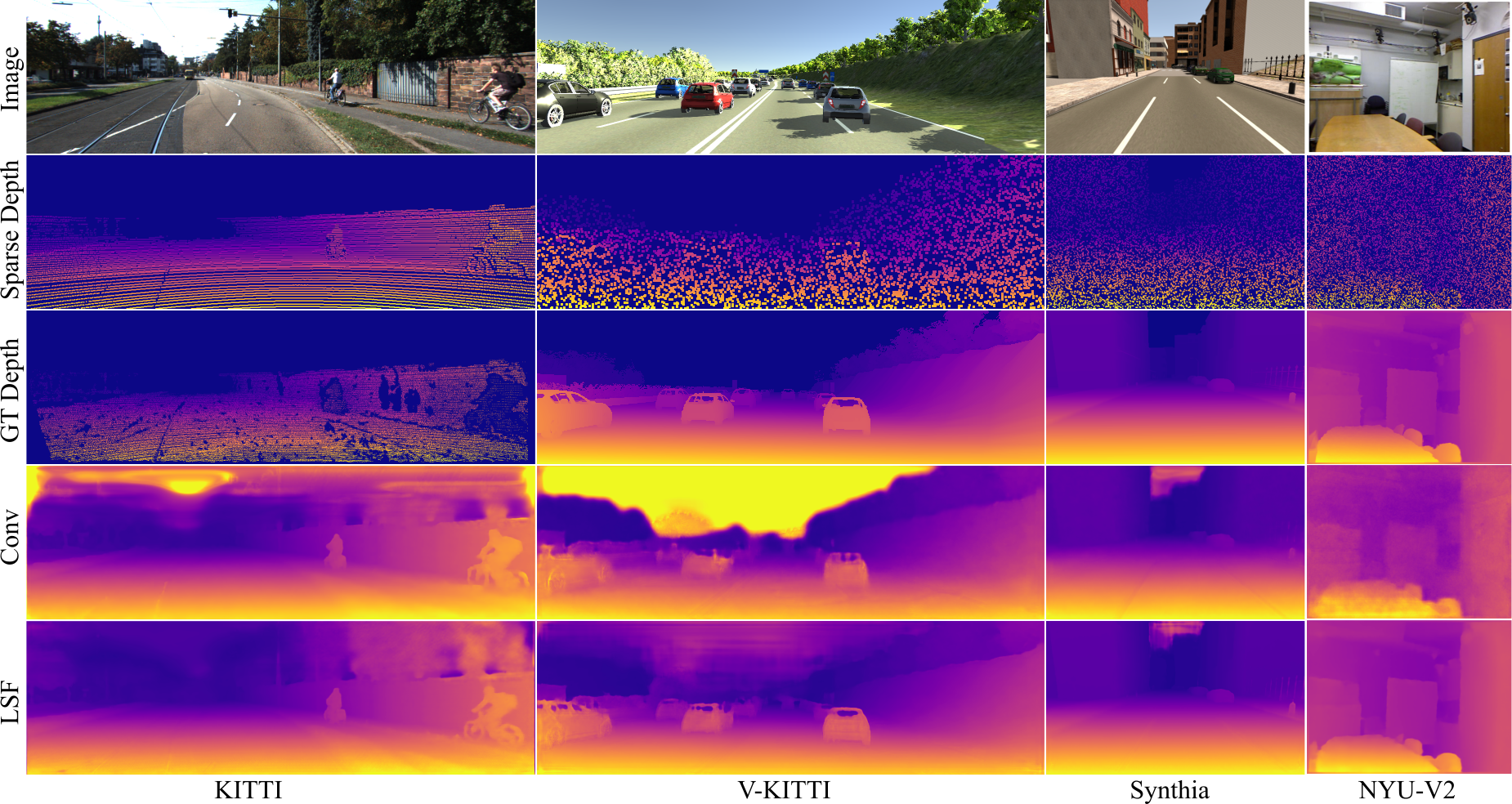}
\end{center}
\caption{Qualitative results of supervised learning on various datasets.
Sparse depths are dilated for visualization purpose (4\% of image resolution).
Artifacts in the upper part of depth prediction from outdoor datasets are due to lack of supervision.}
\label{fig:exp/plots}
\end{figure*}

% ==============================================================================
% \vspace{\baselineskip}
\noindent \textbf{Performance of Linear Fitting.}
Table \ref{tab:exp/perf_sup} shows quantitative comparisons between our proposed linear LSF module from Section \ref{sec:lsf} and the baseline under supervised training.
We see consistent improvements of our linear LSF module over the baseline in all metrics across all datasets.
Note that for \textbf{rgb} input only, the baseline doesn't use any sparse depth information at all.
Thus the large improvement achieved by our fitting method using depth measurements for only 0.2\% of the pixels is quite significant.
For the \textbf{rgbd} case, although the sparse depth map is already used as the input to the baseline network, adding our fitting module better constrains the final prediction to be in accordance with the measurements and
improves the baseline network. 
Since we use the L1 norm as our loss function, the improvement in MAE is bigger than that in RMSE.
Examples of depth prediction are shown in Figure \ref{fig:exp/plots} for qualitative comparisons.

We also perform experiments in which we take a pre-trained baseline method, replace the final convolutional layer with our LSF module and evaluate without re-training. 
This is denoted by \textbf{lsf-}.
Results show that re-training a baseline network with the LSF module allows it to achieve significantly better performance.

Additionally, we compare with PnP \cite{wang2018pnp}, which is a similar method that can be used on many existing networks to improve performance (see Table \ref{tab:exp/perf_sup} and \ref{tab:exp/perf_noise} ).
The main difference is that PnP does not require re-training.
We use the author's official implementation on our baseline network by updating the output of the encoder and run for 5 iterations with update rate 0.01 as suggested in the paper.
We found that although PnP has the advantage no re-training, it takes much longer to run, uses a large amount of memory and yields a smaller improvement compared to ours. Comparisons of runtime are provided in the supplementary material. 

Table \ref{fig:exp/cspn} compares our results to those achieved with  CSPN\cite{Cheng2018DepthEV}. 
The numbers for the CSPN system are taken directly taken from their paper and the official KITTI depth completion benchmark. 
For NYU-V2 we use the same data split they used and sample 500 sparse depths. These results show the improvement afforded by our method.

\begin{table}[h!]
\begin{center}\footnotesize
\begin{tabular}{|c|c|c|c|c|c|c|}
\hline
\multicolumn{2}{|c|}{} &  \multicolumn{2}{|c|}{NYU-V2} & \multicolumn{3}{|c|}{KITTI} \\ \hline
Input & Method  & RMSE & $\delta_1$ & MAE & RMSE & iRMSE  \\ \hline
rgbd & cspn & 0.136 & 99.0 & 0.2795 & 1.0196 & \textbf{2.93} \\
rgbd & lsf2+ & \textbf{0.134} & \textbf{99.3} & \textbf{0.2552} & \textbf{0.8850} & 3.40 \\ \hline
\end{tabular}
\caption{Comparison results on both NYU-V2 and KITTI between CSPN\cite{Cheng2018DepthEV} and our method. 
\textbf{lsf2+} is our LSF module with 2 iterations and Huber loss.
}
\label{fig:exp/cspn}
\end{center}
\end{table}

% ==============================================================================
\begin{figure}[h!]
\begin{center}
\includegraphics[width=\linewidth]{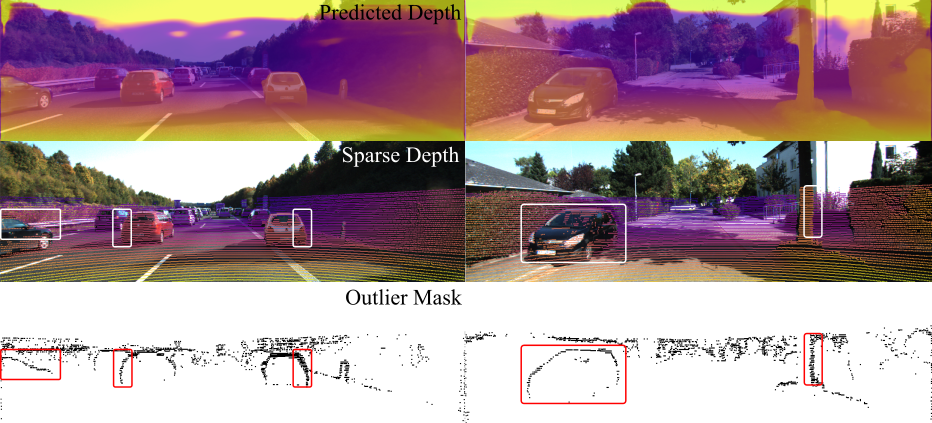}
\end{center}
\caption{When using a robust norm, 
outliers from the input sparse depths can be identified. 
For KITTI dataset, these outliers usually occur at object boundaries, which we highlight a few in rectangles.  
Best view when zoomed in.
}
\label{fig:exp/outlier}
\end{figure}

\noindent \textbf{Dealing with Noise and Outliers.}
To verify the effectiveness of our proposed robustified nonlinear fitting module, 
we inject additive Gaussian noise with a standard deviation of 0.05 meters to sparse depths from NYU-V2, V-KITTI, and Synthia.
We then randomly select 30\% of the available sparse depths to be outliers and set them to random values drawn uniformly from a range between $0.5\times$ to $1.5\times$ of the true depth value.
We left KITTI untouched as it already contains noise and outliers \cite{Cheng2019NoiseAwareUD}.
All nonlinear variants of LSF runs for 2 iterations, which we empirically found to achieve a good balance between performance and efficiency.
We refer the reader to our supplementary material for further discussion on the number of iterations.
We then train various models with different configurations using the corrupted data, which are also grouped by input modalities.
Quantitative results are shown in Table \ref{tab:exp/perf_noise}.

For the \textbf{rgb} case, we ignore the baseline \textbf{conv} as it doesn't use sparse depths and is, therefore, unaffected by noise. 
We again see consistent improvements in all metrics across all datasets.
Notice that for our nonlinear fitting without Huber loss (\textbf{lsf2}), we get worse numbers on some datasets compared to our linear variant (\textbf{lsf}). 
This is because least squares fitting is sensitive to outliers without a robust norm.
There are also some models in the \textbf{rgbd} case where the robustified version (\textbf{lsf2+}) doesn't outperform the linear and nonlinear ones. 
We hypothesize this to be caused by using the corrupted sparse depths as network input which degrades the networks performance early on.
We show in Figure \ref{fig:exp/outlier} that our proposed method is able to identify outliers in the sparse depths and downplay them during fitting.

These results can also be cross-compared with those in Table \ref{tab:exp/perf_sup}, which are all trained on clean data.
Clearly, models trained with clean data outperforms those trained with corrupted ones with the same configuration.
But ours with nonlinear fitting and Huber loss (\textbf{lsf2+}) can sometimes reach similar performance to those trained with clean data even when significant noise and outliers are present.

\begin{figure}[h!]
\begin{center}
\includegraphics[width=\linewidth]{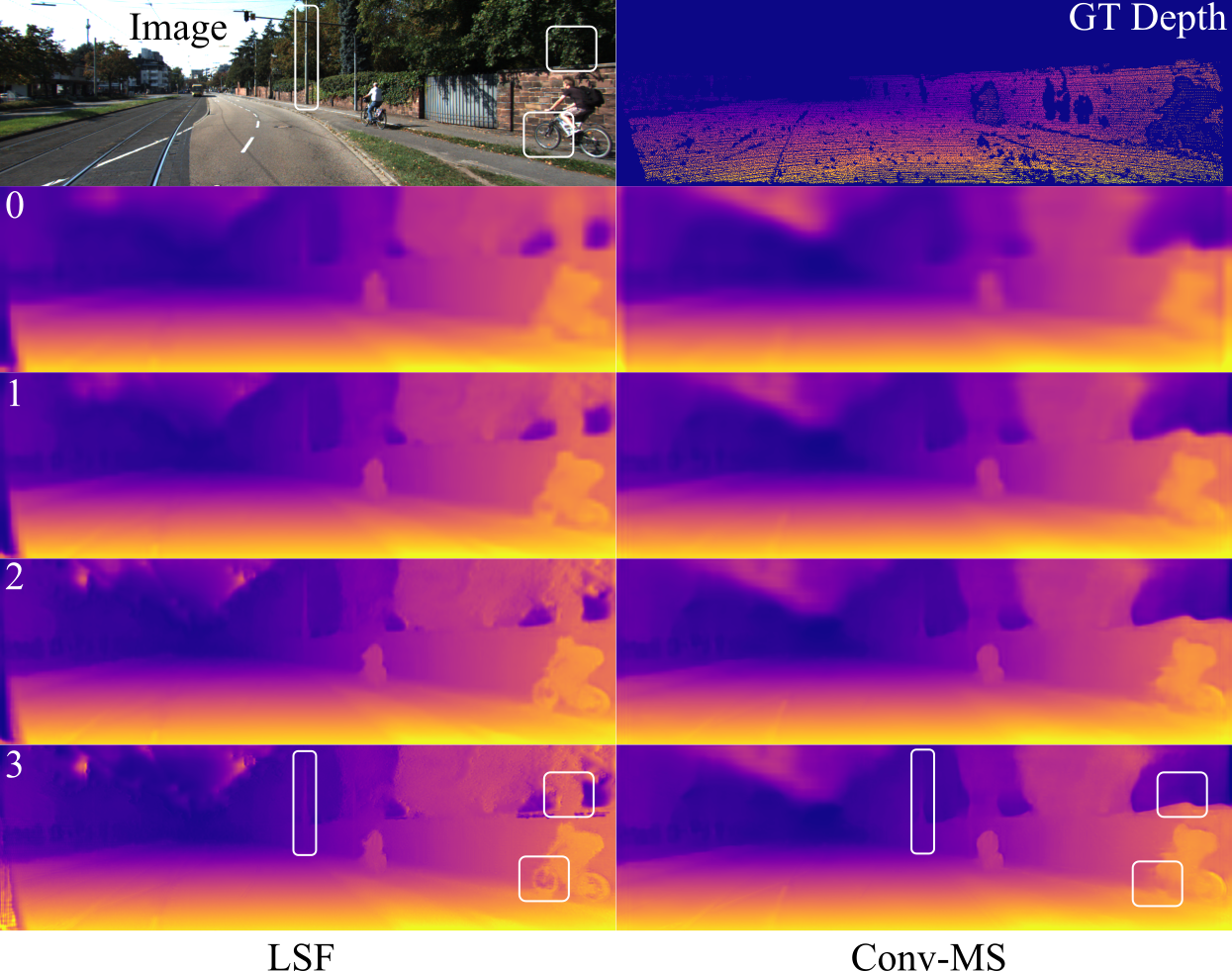}
\end{center}
\caption{Qualitative results of our proposed multi-scale prediction versus the baseline using stereo self-supervision on KITTI dataset. 
All intermediate depth maps are upsampled to the image resolution as suggested in \cite{Godard2018DiggingIS}.
Our multi-scale bases are able to learn a much more detailed depth map compared to the baseline method.
Numbers at top left corner of each image indicate the scale level, where 3 is the full resolution depth map.
Best view when zoomed in.
}
\label{fig:exp/selfsup}
\end{figure}

\vspace{\baselineskip}
\noindent \textbf{Self-supervised Training with Multi-scale Prediction.}
Table \ref{tab:exp/perf_self} shows quantitative comparisons between our linear LSF module with multi-scale basis and the baseline network under both monocular and stereo self-supervised training.
In this case, the baseline network has more parameters because it needs to predict depths at different scales independently.
We again witness consistent improvement in all metrics across all datasets except for $\delta_1$ in KITTI. 
Qualitative results are shown in Figure \ref{fig:exp/selfsup}. 
For all self-supervised training, we use the same hyper-parameters on photometric and smoothness loss as in \cite{Godard2018DiggingIS}, where $\lambda_p = 1.0$ and $\lambda_s=0.001$.
Note in monocular training, we use the ground truth poses directly, as opposed to having a dedicated pose network.

\section{Conclusions}

% In this paper, we propose a least-squares fitting (LSF) module for depth completion task that can be integrated into most existing networks.
% We validate our LSF module on a state-of-the-art depth prediction network with various input modalities (RGB, RGB+SD), training frameworks (supervised, self-supervised), and datasets (KITTI, V-KITTI, Synthia, NYU-V2).
% Our experiments show consistent improvements of our proposed method over the baseline with only slight inference time overhead.
In this paper we propose a novel approach to the depth completion problem that augments deep convolutional networks with a least squares fitting procedure. 
This method allows us to combine some of the best features of modern deep networks and classical regression algorithms. 
% Effectively the scheme splits the regression problem into two sections, the first section of the network produces a basis of depth features which we can view as a low-dimensional manifold that encodes possible scene structures while the second section searches for the point on the manifold that is in best agreement with the available sparse data.
This scheme could be applied to a number of proposed depth completion networks or other regression problems to improve performance. 
Our proposed module is differentiable which means the modified networks can still be trained from end to end. 
This is important because retraining the networks allows them to make better use of the new fitting layer and allows them to produce better depth bases from the input data. 
% For example our experiments show that retraining an rgb only network with the new fitting layer allows it to produce significantly better results than simply adding the fitting scheme without retraining the network.
We then describe how a linear least squares fitting scheme could be extended to incorporate robust estimation to improve resilience to noise and outliers which are common in real world data. 
We also show the method can be employed in self-supervised settings where no ground truth is available.
% Here we suggest a novel scheme to efficiently produce the multi-scale representations of depth that are required for modern training methods that minimize the photometric error between warped versions of the original images.
We validate our fitting module on a state-of-the-art depth completion network with various input modalities, training frameworks, and datasets.
% Our experiments show consistent and significant improvements of our proposed method over the baseline with only slight inference time overhead.

% novel adaptive fitting stage

% combines best features of deep learning and fitting

% Can be extended to incorporate m estimation to improve resilience to noise and outliers at the cost of extra computation

% unlike previously proposed approaches, this method only adds modestly to the computational cost.

% Approach can be employed in self-supervised setting where no depth data is available where we suggest a novel scheme to efficiently produce multi-scale representations of depth.

% Improvement in training time and stability over baseline method

% Retraining indicates that with this the network learns a stronger basis as evidenced by table 2

% evaluate it by showing that it significantly improves the performance of a strong baseline model on multiple datasets.

% Can readily be incorporated into other depth completion networks. Could be used for other regression tasks.

% Greater resilience to sparse data

\section{Acknowledgement}

We would like to acknowledge the support of Novateur Research Solutions and an Nvidia NVAIL grant.

\section{Ablation Study}

\begin{table*}[h!]
\begin{center}\footnotesize
\begin{tabular}{|c|c|c|c|c|c|c|c|c|c|c|c|}
\hline
\multicolumn{3}{|c|}{Trained on KITTI} & \multicolumn{3}{c|}{NYU-V2}& \multicolumn{3}{c|}{V-KITTI}  & \multicolumn{3}{c|}{Synthia} \\ \hline
Input & Method & Sparse & MAE & RMSE & $\delta_1$ & MAE & RMSE & $\delta_1$ & MAE & RMSE & $\delta_1$ \\ \hline\hline
rgbd & conv & 4\% & 0.5318 & 0.8670 & 67.93 & 2.8855 & 9.1813 & 90.02 & 4.7380 & 14.408 & 69.51 \\
rgbd & lsf & 4\% & \textbf{0.2590} & \textbf{0.8155} & \textbf{90.59} & \textbf{1.9189} & \textbf{6.9789} & \textbf{92.80} & \textbf{3.1198} & \textbf{9.5432} & \textbf{83.07} \\ \hline
\multicolumn{3}{|c|}{(conv-lsf)/lsf} & +51\% &  +6\% &  &  +33\% & +24\% & & +34\% & +34\%   & \\ \hline
\end{tabular}
\end{center}
\caption{Quantitative results of supervised training on KITTI and evaluate on other datasets.
Here \textbf{conv} denotes the s2d baseline \cite{Ma2018SelfsupervisedSS},
\textbf{lsf} is our linear LSF module.
Percentage values listed under the \textbf{Sparse} column indicates sparsity of image resolution (around 20k for KITTI).
Best results in each category are in \textbf{bold}.}
\label{tab:general}
\end{table*}

\subsection{Generalization}
We demonstrate the generalization capability of our proposed module compared to the s2d baseline. 
We train both models on the KITTI depth completion dataset and evaluate on the rest (NYU-V2, V-KITTI, and Synthia). 
Table \ref{tab:general} shows the quantitative results of this set of experiments.
We change the evaluation strategy of Synthia and V-KITTI to cap at 80 meters since the maximum depth of KITTI is around 85 meters, but the maximum input sparse depth is still 130 meters for both datasets. 
We observe similar improvements in networks with our LSF module, which shows that it is able to generalize well to other datasets.

\subsection{Convergence Rate}
\begin{figure}[h!]
\begin{center}
\includegraphics[width=\linewidth]{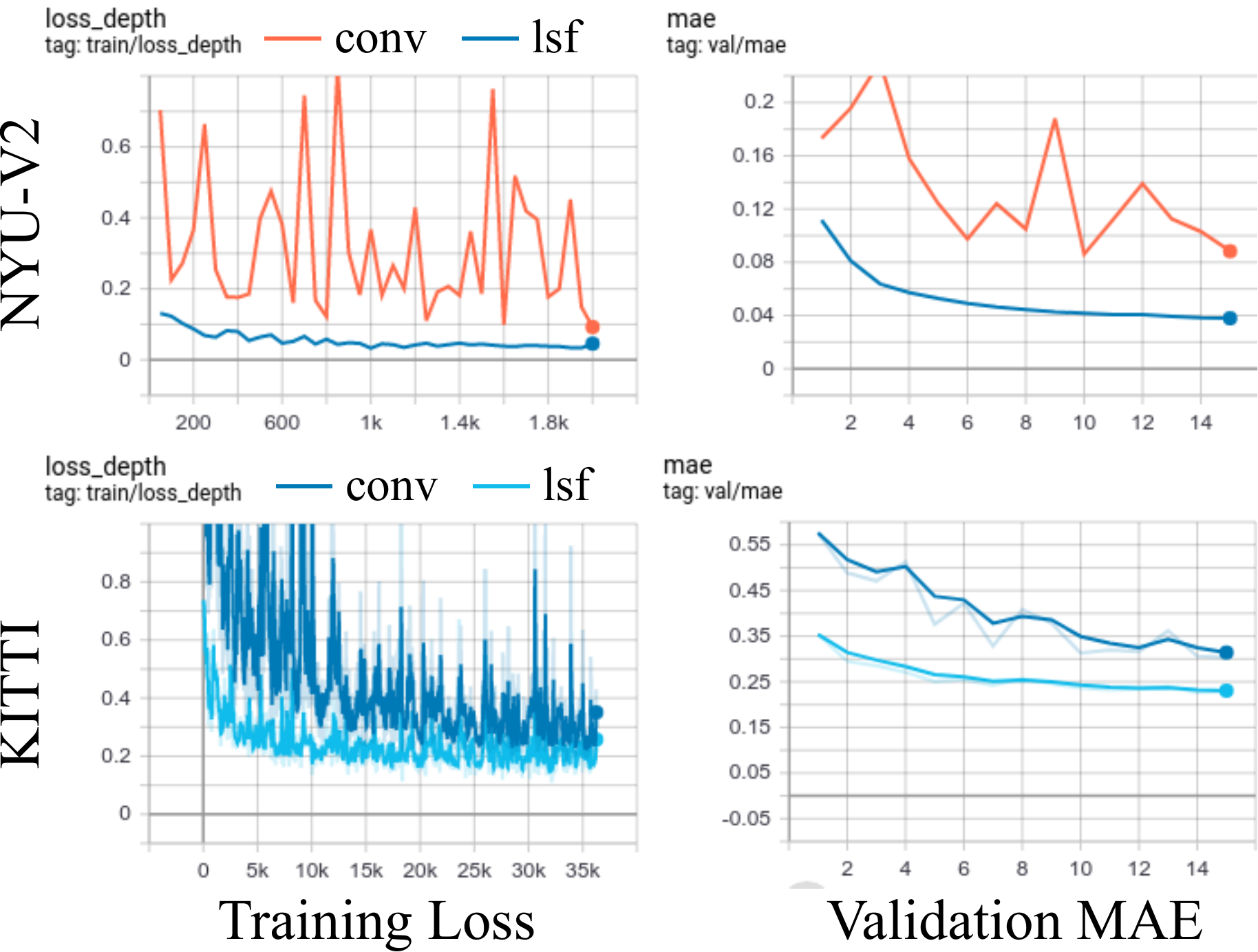}
\end{center}
\caption{Tensorboard records of training loss (L1) and validation MAE on NYU-V2 and KITTI dataset using both color image and sparse depth as input. 
Left column shows the training loss \wrt training iterations and right column shows validation MAE \wrt training epochs.}
\label{fig:exp/tensorboard}
\end{figure}

Figure \ref{fig:exp/tensorboard} shows snapshots of Tensorboard\footnote{\url{https://www.tensorflow.org/guide/summaries_and_tensorboard}} records of several training sessions of the s2d baseline with and without our LSF module. 
Notice that on both NYU-V2 and KITTI datasets (and others that are not shown here), training with our LSF module has a faster convergence rate, a more stable learning curve and as a result, better validation performance.
This trend is observed in all our experiments with various datasets and baseline networks \cite{Shivakumar2019DFuseNetDF, Gansbeke2019SparseAN, Ma2018SelfsupervisedSS}.
Given this property, we hypothesize that by using our LSF module, we can quickly fine-tune a pre-trained baseline model to another dataset.

\subsection{Multi-scale Bases}
We've already shown in the main paper that our least squares fitting with multi-scale bases outperforms the baseline in various self-supervised training frameworks. 
Here, we show additional experiments of multi-scale vs single-scale bases in supervised learning, using a multi-scale bases of size (4, 8, 16, 32) versus a single scale basis of size 60.
These two schemes have exactly the same same parameters.
The results in table \ref{tab:exp/multi_vs_single} show that using the multi-scale formulation is even beneficial in settings where only a single full resolution depth map is needed.
This improvement can be partially explained by the fact that gradients can now directly flow back to the intermediate decoder layers rather than indirectly from the final layer. 
This is also related to the idea of deep supervision \cite{Wang2015TrainingDC}, which demonstrated that adding intermediate supervision facilitates training of very deep neural networks.
Therefore, we use multi-scale bases in all our experiment with the baseline, 
for it doesn't introduce extra parameters, yields superior performance and is compatible with both supervised and self-supervised training paradigms.
We use single-scale bases for FusionNet \cite{Gansbeke2019SparseAN} because their network is not designed for multi-scale prediction.

\begin{table}[h!]
\begin{center}\footnotesize
\begin{tabular}{|c|c|c|c|c|c|c|}
\hline
 & \multicolumn{3}{c|}{NYU-V2}& \multicolumn{3}{c|}{Synthia} \\ \hline
Basis & MAE & RMSE & $\delta_1$ & MAE & RMSE & $\delta_1$ \\ \hline
60 & 0.0315 & 0.0757 & \textbf{99.83} & 0.5332 & 3.1353 & 97.83 \\
4,8,16,32 & \textbf{0.0300} & \textbf{0.0735} & \textbf{99.83} & \textbf{0.5317} & \textbf{3.1057} & \textbf{97.84} \\ \hline
\end{tabular}
\end{center}
\caption{Quantitative results of single-scale vs multi-scale bases on various datasets under supervised training using s2d baseline network. 
All networks are trained using the LSF module with rgbd input where the sparse depth has a 4\% sparsity.
Best results in each category are in \textbf{bold}.}
\label{tab:exp/multi_vs_single}
\end{table}

\subsection{Underdetermined Systems}
In our setup, the linear system will become underdetermined when the number of sparse samples is smaller than the number of basis.
Although this has rarely been evaluated in prior literature (100 samples in \cite{wang2018pnp}), we test our module under such extreme cases (50 samples).

We use 50 sparse depths samples for both training and validation, which is less than the dimension of the basis used (60) which makes the linear system in the LSF module underdetermined. 
Due to the small number of samples, we increase the regularization parameter $\lambda$ to 0.01.
Note that PnP \cite{wang2018pnp} does not require training and operates directly on the pretrained baseline.
Table \ref{tab:underdet} shows the result of this experiment. 
In this case, our LSF module does not outperform the baseline nor PnP, but it still provides a reasonable result due to the regularization in the fitting.

When the number of sparse depth samples goes to 0, our module will inevitably fail, while the other baselines are still able to output a solution.
This is indeed a drawback to our method and we plan to address this issue in future work.

\begin{table}[h!]
\begin{center}\footnotesize
\begin{tabular}{|c|c|c|c|c|c|c|}
\hline
 & \multicolumn{3}{c|}{NYU-V2} & \multicolumn{3}{c|}{V-KITTI} \\ \hline
Method & MAE& RMSE& $\delta_1$& MAE& RMSE& $\delta_1$ \\ \hline
conv & \textbf{0.2218} & \textbf{0.4170} & 92.03 & 6.1841 & 14.273 & 74.13 \\
pnp & 0.2233 & 0.4170 & \textbf{92.12} & \textbf{6.0465} & \textbf{14.119} & \textbf{75.21}\\ 
lsf  & 0.3313 & 0.5464 & 83.63 & 8.2031 & 16.686 & 55.75\\ 
\hline
\end{tabular}
\end{center}
\caption{Quantitative results of training and testing on 50 depth samples.
All networks take rgbd input. 
Here \textbf{conv} denotes the s2d baseline, \textbf{pnp} denotes applying PnP \cite{wang2018pnp} module on the trained \textbf{conv} network with 5 iterations, \textbf{lsf} is our linear fitting module.
Best results in each category are in \textbf{bold}.}
\label{tab:underdet}
\end{table}

\subsection{Number of Iterations for Nonlinear Fitting}
We ran several experiments training the baseline network with our nonlinear LSF module while varying the number of iterations used.
Results are shown in table \ref{tab:exp/num_iters}.
Like many iterative approaches, we see a diminishing return with increasing number of iterations.
Empirically, we found that 2 iterations strike a balance between performance and efficiency, and thus use it across all our nonlinear fitting experiments.
However, we did not observe any instability with more iterations, other than a marginal variation in validation metrics.

\begin{table}[h!]
\begin{center}\footnotesize
\begin{tabular}{|c|c|c|c|c|c|c|}
\hline
 & \multicolumn{3}{c|}{NYU-V2} & \multicolumn{3}{c|}{V-KITTI} \\ \hline
Method & MAE& RMSE& $\delta_1$& MAE& RMSE& $\delta_1$ \\ \hline
conv  & 0.1089  & 0.1679 & 99.20 & 1.5683 & 4.8982 & 94.68 \\
lsf0  & 0.0300  & 0.0735 & \textbf{99.83} & \textbf{1.2598} & 4.6227 & \textbf{97.32} \\ 
lsf1  & 0.0293 & 0.0721 & \textbf{99.83} & 1.2932 & \textbf{4.5717} &  96.92 \\ 
lsf2  & 0.0293 & \textbf{0.0720} & \textbf{99.83} & 1.2643 & 4.6114 & 97.07 \\ 
lsf3  & \textbf{0.0292} & \textbf{0.0720} & \textbf{99.83} & 1.3047 & 4.6159 & 96.78 \\ 
\hline
\end{tabular}
\end{center}
\caption{Quantitative results of training our LSF module with various numbers of iterations using the baseline. 
Here \textbf{conv} denotes the baseline, \textbf{lsf} is our linear fitting module with the trailing digit indicating the number of iterations. For example, \textbf{lsf0} means linear fitting, \textbf{lsf2} means nonlinear fitting with 2 iterations.
Best results in each category are in \textbf{bold}.}
\label{tab:exp/num_iters}
\end{table}

\subsection{Runtime Comparison}

\begin{figure}[h!]
\begin{center}
\includegraphics[width=\linewidth]{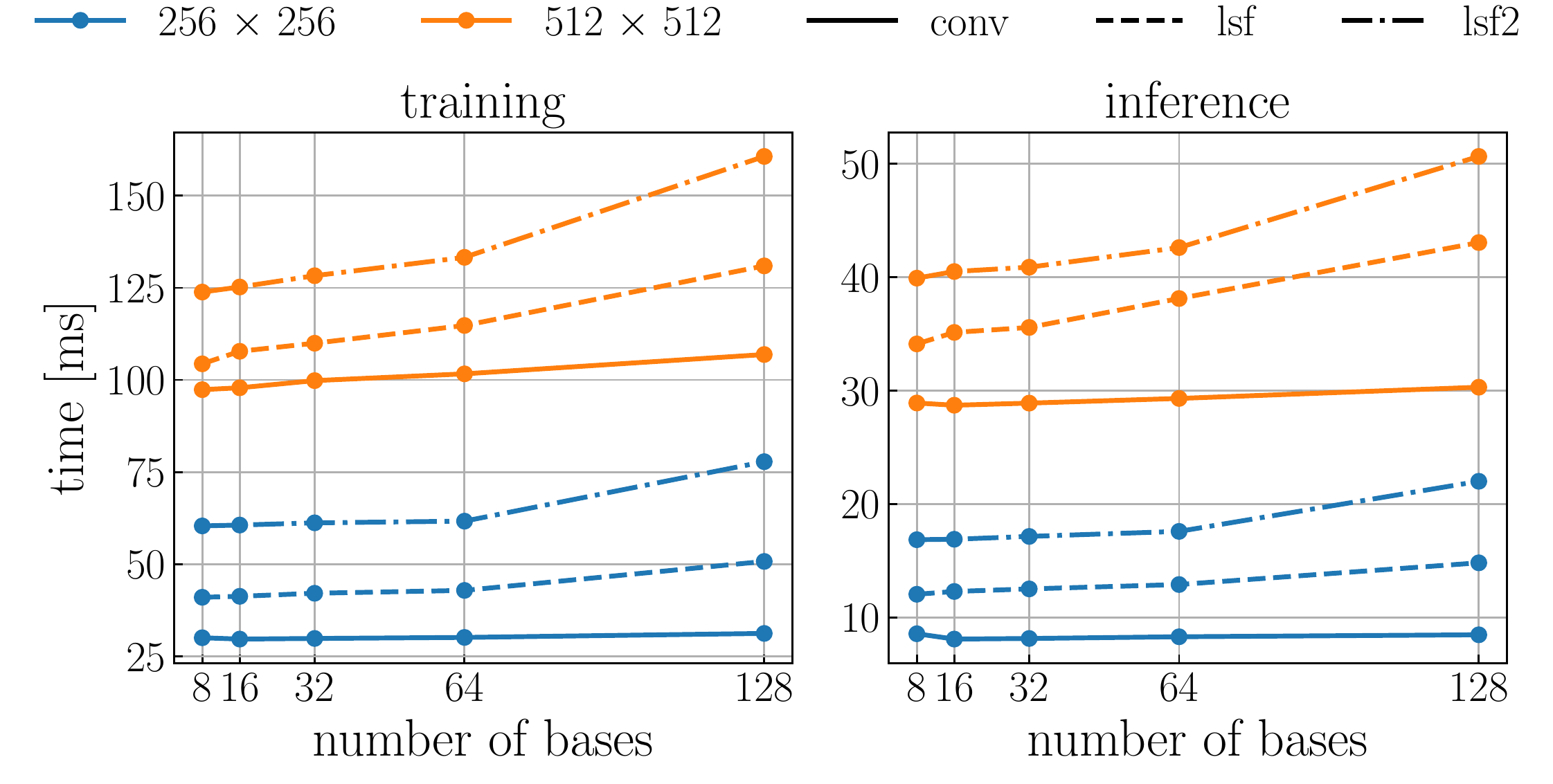}
\end{center}
\caption{Training and inference time of our LSF module compared to the s2d baseline on one sample for one iteration on a single GTX 1080-Ti GPU.
Here, \textbf{conv} refers to the baseline with a final convolution layer, \textbf{lsf} refers to our module with linear fitting, and \textbf{lsf2} is the nonlinear version with 2 Gauss-Newton steps.
Different colors indicate different image sizes.}
\label{fig:exp/runtime}
\end{figure}

In Figure \ref{fig:exp/runtime}, we show runtime comparisons between variants of our LSF modules and the baseline in both training and inference.
The increase in computation time is due to the (repeated) solving of a linear system of equations whose complexity depends on the size of the basis.
The number of sparse depth samples has a very small impact to the runtime (as explained above), and we fix it to 1024 in this experiment.
Our linear LSF module adds on average 46\% to inference time compared to the baseline network. 
Note that the times provided in the graph represent the total time required for a complete forward/backward pass through the network.

\begin{table}[h!]
\begin{center}
\begin{tabular}{|c|c|c|c|c|c|}
\hline
& conv & lsf & lsf2 & cspn & pnp \\ \hline
Time [ms] & 34.4 & 42.3 & 45.9 & 53.9 & 335.2 \\ \hline
\end{tabular}
\end{center}
\caption{
Runtime comparison across different methods on a single 512$\times$512 image with 1024 sparse depth samples during inference.
Add methods are using the baseline network. 
We run \textbf{cspn} for 12 iterations and \textbf{pnp} for 5 iterations as suggested in their paper.
}
\label{tab:exp/runtime}
\end{table}

\section{More Experiments}

\begin{table*}[h!]
\begin{center}\footnotesize
\begin{tabular}{|c|c|c|c|c|c|c|c|c|c|c|c|c|c|c|}
\hline
\multicolumn{3}{|c|}{Supervised Training} & \multicolumn{3}{c|}{NYU-V2} & \multicolumn{3}{c|}{V-KITTI} & \multicolumn{3}{c|}{Synthia} & \multicolumn{3}{c|}{KITTI} \\ \hline
Input & Method & Sparse & MAE & RMSE & $\delta_1$ & MAE & RMSE & $\delta_1$ & MAE & RMSE & $\delta_1$ & MAE & RMSE & $\delta_1$ \\ \hline\hline
rgbd & conv & 4\% & 0.1035 & 0.1454 & 99.00 & 2.3531 & 5.5823 & 86.28 & 0.8052 & 3.1054 & 95.87 & 0.2790 & 1.0001 & 99.58 \\
rgbd & lsf & 4\% & \textbf{0.0338} & \textbf{0.0752} & \textbf{99.82} & \textbf{1.4440} & \textbf{4.5085} & \textbf{95.83} & \textbf{0.6754} & \textbf{2.9411} & \textbf{96.72} & \textbf{0.2707} & \textbf{0.9142} & \textbf{99.68} \\ \hline
\multicolumn{3}{|c|}{(conv - lsf)/ conv} & +67\% & +48\% & & +39\% & 19\% & & +16\% & +5\% & & 3\% & +9\% & \\ \hline
\end{tabular}
\end{center}
\caption{Quantitative results of supervised training on various datasets using FusionNet\cite{Gansbeke2019SparseAN} baseline .
Here \textbf{conv} denotes the baseline with final convolution,
\textbf{lsf} is our linear LSF module.
Percentage values listed under the \textbf{Sparse} column indicates sparsity of image resolution.
Best results in each category are in \textbf{bold}.}
\label{tab:exp/perf_fusion}
\end{table*}

\begin{table*}[h!]
\begin{center}\footnotesize
\begin{tabular}{|c|c|c|c|c|c|c|c|c|c|c|c|c|c|c|}
\hline
\multicolumn{3}{|c|}{Supervised Training} & \multicolumn{3}{c|}{NYU-V2} & \multicolumn{3}{c|}{V-KITTI} & \multicolumn{3}{c|}{Synthia} & \multicolumn{3}{c|}{KITTI} \\ \hline
Input & Method & Sparse & MAE & RMSE & $\delta_1$ & MAE & RMSE & $\delta_1$ & MAE & RMSE & $\delta_1$ & MAE & RMSE & $\delta_1$ \\ \hline\hline
rgbd & conv & 4\% & 0.1217 & 0.2198 & 97.57 & 3.5029 & 8.2014 & 83.05 & 1.5143 & 4.8709 & 90.19 & 0.7732 & 1.9537 & 98.21 \\
rgbd & lsf & 4\% & \textbf{0.0722} & \textbf{0.1517} & \textbf{99.18} & \textbf{2.6840} & \textbf{6.6656} & \textbf{90.89} & \textbf{1.2929} & \textbf{4.2834} & \textbf{92.40} & \textbf{0.5526} & \textbf{1.6380} & \textbf{98.60} \\ \hline
\multicolumn{3}{|c|}{(conv - lsf)/ conv} & +41\% & +31\% & & +23\% & 19\% & & +14\% & +13\% & & 29\% & +16\% & \\ \hline
\end{tabular}
\end{center}
\caption{Quantitative results of supervised training on various datasets using DFuseNet\cite{Shivakumar2019DFuseNetDF} baseline .
Here \textbf{conv} denotes the baseline with final convolution,
\textbf{lsf} is our linear LSF module.
Percentage values listed under the \textbf{Sparse} column indicates sparsity of image resolution.
Best results in each category are in \textbf{bold}.}
\label{tab:exp/perf_dfuse}
\end{table*}

% To show that our module is applicable to other networks, we conduct extra experiments against other networks that uses completely different fusion strategies and network architectures.

\subsection{FusionNet}

FusionNet \cite{Gansbeke2019SparseAN} generates dense depth predictions by combining both global and local information guided by color images.
Their network also learns two uncertainty maps that fuses the global and local depth maps.
The global branch generates a global depth prediction with uncertainty as well as a guidance map, which is then used in the local branch to predict a local depth map with uncertainty.
These two depth maps are then linearly combined with normalized weights from the corresponding uncertainty maps.
In terms of architectural difference, their network uses an ERFNet\cite{Romera2018ERFNetER} in the global branch and two hourglass networks in the local branch. 
ERFNet is a network designed for efficient semantic segmentation and has around 3M parameters (while ResNet18 has around 15M).
There is no multi-scale bases in this baseline and the final activation function is ReLU.

We use the network implementation from the official repository\footnote{\url{https://github.com/wvangansbeke/Sparse-Depth-Completion}}.
We make the following modifications to their network so that our LSF module can be attached:
1) instead of using the uncertainty maps to weight the predicted depth, we use them to weight the penultimate feature maps (bases). 
Because of the linearity of the convolution operations, this is equivalent to the original implementation.
2) we trained the network from scratch as a whole instead of using pretrained weights on CityScapes\cite{Cordts2016Cityscapes} and break the training into two steps (first global then local).
We tried our best to follow the training settings in the original paper with a starting learning rate of 1e-3 and L2 loss instead of L1. 
All networks are trained for 15 epochs.

Quantitative results are show in table \ref{tab:exp/perf_fusion}.
Note that some of the results might differ from the reported numbers from their paper, which can be attributed to many factors such as differing random seeds, training epochs and weight initialization.
However, we made an honest effort to make sure that the network architecture was the same as the original and we believe that the improvements in performance offered by our method are representative.

\subsection{DFuseNet}

The main baseline \cite{Ma2018SelfsupervisedSS} that we used adopts an early fusion strategy to combine color and depth information, where the two streams of information are combined after the first convolution layer.
In DFuseNet\cite{Shivakumar2019DFuseNetDF}, the authors instead favor a late fusion approach and use a Spatial Pyramid Pooling (SPP) \cite{He2014SpatialPP} block in each branch to incorporate more contextual information. In terms of architectural differences, their network does not use skip connection and is trained from scratch. 

We use the network implementation from the official repository\footnote{\url{https://github.com/ShreyasSkandanS/DFuseNet}}, but make the following modifications for more stable training: 
1) add a batch normalization \cite{Ioffe2015BatchNA} layer after every convolution except for the last one, 
2) remove one scale (\texttt{pool64}) from the SPP block to make the network trainable with a reasonable batch size on our GPU,
3) change the decoder output from (64, 32, 16, 1) channels to (4, 8, 16, 32) channels bases.
The rest of the training parameters are kept the same as described in their paper.

Quantitative results are shown in Table \ref{tab:exp/perf_dfuse}. Our LSF module again improve the baseline by a significant amount under the same training setting.

% \subsection{KITTI Depth Completion Benchmark}

% \begin{figure*}[h!]
% \begin{center}
% \includegraphics[width=\linewidth]{diagrams/kitti.png}
% \end{center}
% \caption{Replace with official results on KITTI benchmark, pending}
% \label{fig:exp/kitti}
% \end{figure*}

% Figure \ref{fig:exp/kitti} shows more qualitative results on the KITTI depth completion dataset. We train two networks, one with s2d baseline and one LSF with 2 iterations and Huber loss.
% Both are trained with depth and stereo supervision. 

{\small
\bibliographystyle{ieee}
\bibliography{root}
}

\end{document}